\documentclass[letterpaper, 10 pt, conference]{ieeeconf}  

\IEEEoverridecommandlockouts                              

\usepackage[left=1.88cm,right=1.88cm,top=1.88cm,bottom=1.88cm]{geometry}
\usepackage{amsmath,amssymb,amsfonts}
\usepackage{graphicx}
\usepackage{textcomp}
\usepackage{xcolor}
\usepackage{bbm}
\let\labelindent\relax
\usepackage{enumitem}
\usepackage{tabularx}

\usepackage{times}
\usepackage{multicol}

\usepackage{graphicx}
\usepackage{bm}
\usepackage[nolist]{acronym}

\usepackage[linesnumbered,ruled]{algorithm2e}

\usepackage{mathtools}
\usepackage{array}
\usepackage{dblfloatfix}
\usepackage{subfigure}
\usepackage{diagbox}
\usepackage{etoolbox}\AtBeginEnvironment{algorithmic}{\small}
\usepackage{color,soul} 

\usepackage{lipsum}
\usepackage[colorinlistoftodos]{todonotes}

\usepackage{amsthm}

\usepackage{bbm} 
\usepackage[colorlinks = True, linkcolor = blue, citecolor = blue]{hyperref}

\usepackage{cleveref}
\usepackage{rotating}
\usepackage{multirow}
\usepackage{caption}
\usepackage[
    style=ieee,
    doi=false,
    isbn=false,
    url=false,
    eprint=false,
    backend=biber,
    natbib=true
    ]{biblatex}

\bibliography{refs}

\title{\LARGE \bf
Selecting Spots by Explicitly Predicting Intention from Motion History Improves Performance in Autonomous Parking
 }

\author{Long Kiu Chung$^{1,2}$, David Isele$^1$, Faizan M. Tariq$^1$, Sangjae Bae$^1$, Shreyas Kousik$^2$, and Jovin D'sa$^1$
\thanks{
$^1$Honda Research Institute (HRI), Mountain View, CA.
$^2$ Georgia Institute of Technology, Atlanta, GA.
Corresponding Authors: \href{mailto:lchung33@gatech.edu}{\texttt{lchung33@gatech.edu}}, \href{mailto:jovin_dsa@honda-ri.com}{\texttt{jovin\_dsa@honda-ri.com}}. 
All work was done when Long Kiu Chung was employed by HRI.
}}

\begin{document}


\newif\ifshowcomments
\showcommentsfalse 

\ifshowcomments
    \newcommand{\bae}[1]{\hl{[SB: #1]}\protect\color{black}} 
    \newcommand{\di}[1]{\hl{[DI: #1]}\protect\color{black}} 
    \newcommand{\ft}[1]{\hl{[FT: #1]}\protect\color{black}} 
    \newcommand{\jd}[1]{\hl{[JD: #1]}\protect\color{black}} 
    \newcommand{\edgar}[1]{{\color{red}{LKC: #1}}\protect\color{black}}
    \newcommand{\shrey}[1]{\hl{[SK: #1]}\protect\color{black}}
    
\else
    \newcommand{\bae}[1]{}
    \newcommand{\di}[1]{}
    \newcommand{\ft}[1]{}
    \newcommand{\jd}[1]{}
    \newcommand{\edgar}[1]{}
    \newcommand{\shrey}[1]{}
\fi

\newtheorem{defn}{Definition}
\newtheorem{rem}[defn]{Remark}
\newtheorem{lem}[defn]{Lemma}
\newtheorem{prop}[defn]{Proposition}
\newtheorem{assum}[defn]{Assumption}
\newtheorem{ex}[defn]{Example}
\newtheorem{runex}[defn]{Running Example}
\newtheorem{thm}[defn]{Theorem}
\newtheorem{cor}[defn]{Corollary}
\newtheorem{problem}[defn]{Problem}

\providecommand{\R}{\ensuremath \mathbb{R}}
\newcommand{\Rpos}{\R_{>0}}
\newcommand{\Rnonneg}{\R_{\geq0}}
\newcommand{\N}{\ensuremath \mathbb{N}}

\newcommand{\stateset}{S}
\newcommand{\pointset}{P}
\newcommand{\lotset}{L}
\newcommand{\roadcenter}{C}
\newcommand{\sensor}{Q}
\newcommand{\bezcurve}{C_{\bezlbl}}
\newcommand{\explore}{E}

\newcommand{\regtext}[1]{\mathrm{\textnormal{#1}}}
\newcommand{\vc}[1]{\mathbf{#1}}

\newcommand{\tp}{^\intercal}
\newcommand{\norm}[1]{\left\Vert#1\right\Vert}
\newcommand{\abs}[1]{\left\vert#1\right\vert}
\newcommand{\indicator}{\mathbbm{1}}
\newcommand{\atantwo}{\regtext{atan2}}
\newcommand{\cardinality}{\regtext{n}}
\newcommand{\sign}{\regtext{sign}}
\newcommand{\interior}{\regtext{int}}
\newcommand{\exterior}{\regtext{ext}}

\newcommand{\zeros}{\mathbf{0}}
\newcommand{\ones}{\mathbf{1}}
\newcommand{\state}{\vc{s}}
\newcommand{\ctrl}{\vc{u}}
\newcommand{\point}{\vc{p}}
\newcommand{\coef}{\vc{c}}

\newcommand{\otherlbl}{\regtext{other}}
\newcommand{\spotlbl}{\regtext{spot}}
\newcommand{\pedlbl}{\regtext{ped}}
\newcommand{\roadlbl}{\regtext{road}}
\newcommand{\rstartlbl}{\regtext{start}}
\newcommand{\rendlbl}{\regtext{end}}
\newcommand{\tangentlbl}{\regtext{T}}
\newcommand{\normallbl}{\regtext{N}}
\newcommand{\passivelbl}{\regtext{pass}}
\newcommand{\vaclbl}{\regtext{vac}}
\newcommand{\occlbl}{\regtext{occ}}
\newcommand{\statvehlbl}{\regtext{sv}}
\newcommand{\dynvehlbl}{\regtext{dv}}
\newcommand{\dynpedlbl}{\regtext{dp}}
\newcommand{\parklbl}{\regtext{park}}
\newcommand{\irptlbl}{\regtext{irpt}}
\newcommand{\steallbl}{\regtext{stl}}
\newcommand{\intentlbl}{\regtext{int}}
\newcommand{\entlbl}{\regtext{ent}}
\newcommand{\lotlbl}{\regtext{lot}}
\newcommand{\canlbl}{\regtext{can}}
\newcommand{\bevlbl}{\regtext{BEV}}
\newcommand{\deflbl}{\regtext{def}}
\newcommand{\predlbl}{\regtext{pred}}
\newcommand{\bezlbl}{\regtext{bez}}
\newcommand{\obslbl}{\regtext{obs}}
\newcommand{\trajlbl}{\regtext{traj}}
\newcommand{\explbl}{\regtext{exp}}
\newcommand{\goallbl}{\regtext{goal}}
\newcommand{\smoothlbl}{\regtext{smooth}}
\newcommand{\backlbl}{\regtext{back}}
\newcommand{\planlbl}{\regtext{plan}}
\newcommand{\raylbl}{\regtext{ray}}

\newcommand{\veh}{V}
\newcommand{\spot}{S}
\newcommand{\ped}{P}
\newcommand{\road}{R}
\newcommand{\agent}{A}
\newcommand{\ray}{B}

\newcommand{\xs}{x}
\newcommand{\ys}{y}
\newcommand{\thetas}{\theta}
\newcommand{\vs}{v}
\newcommand{\omegas}{\omega}
\newcommand{\gammas}{\gamma}

\newcommand{\ts}{t}
\newcommand{\dt}{\Delta \ts}
\newcommand{\tf}{\ts_{\regtext{f}}}
\newcommand{\tpast}{\ts_{\regtext{hist}}}
\newcommand{\taus}{\tau}
\newcommand{\tpark}{\ts_{\parklbl}}
\newcommand{\tpred}{\ts_{\predlbl}}

\newcommand{\ndim}{n}
\newcommand{\nother}{\ndim_{\otherlbl}}
\newcommand{\nspot}{\ndim_{\spotlbl}}
\newcommand{\nray}{\ndim_{\raylbl}}
\newcommand{\nped}{\ndim_{\pedlbl}}
\newcommand{\nroad}{\ndim_{\roadlbl}}
\newcommand{\nvac}{\ndim_{\vaclbl}}
\newcommand{\nocc}{\ndim_{\occlbl}}
\newcommand{\nstatveh}{\ndim_{\statvehlbl}}
\newcommand{\ndynveh}{\ndim_{\dynvehlbl}}
\newcommand{\ndynped}{\ndim_{\dynpedlbl}}
\newcommand{\nirpt}{\ndim_{\irptlbl}}
\newcommand{\nsteal}{\ndim_{\steallbl}}

\newcommand{\length}{\ell}
\newcommand{\width}{w}
\newcommand{\radius}{r}

\newcommand{\tplan}{\ts_{\planlbl}}
\newcommand{\ctrlplan}{\tilde{\ctrl}}
\newcommand{\pointsetplan}{\tilde{\pointset}}

\newcommand{\spotset}{I}
\newcommand{\vehset}{K}
\newcommand{\pedset}{M}

\newcommand{\is}{i}
\newcommand{\js}{j}
\newcommand{\ks}{k}
\newcommand{\ms}{m}
\newcommand{\qs}{q}
\newcommand{\isplan}{\tilde{\is}}

\newcommand{\parkpredictint}{\vc{f}_{\regtext{PP+}}}
\newcommand{\ppweight}{\vc{g}_{\regtext{PP+}}}
\newcommand{\parkpredicttraj}{\vc{f}_{\regtext{PP+}_{\trajlbl}}}
\newcommand{\score}{\xi}
\newcommand{\prob}{\eta}
\newcommand{\img}{\vc{V}}
\newcommand{\dist}{d}
\newcommand{\ang}{a}

\newcommand{\belief}{b}
\newcommand{\intbelief}{\tilde{\belief}}
\newcommand{\beliefmap}{\vc{b}}
\newcommand{\intbeliefmap}{\tilde{\beliefmap}}
\newcommand{\beliefthresh}{\beta}
\newcommand{\goalthresh}{\mu}
\newcommand{\egothresh}{\delta}
\newcommand{\predstate}{\bar{\state}}
\newcommand{\predpoint}{\bar{\point}}
\newcommand{\predang}{\bar{\thetas}}
\newcommand{\predvel}{\bar{\vs}}
\newcommand{\predpoly}{\bar{\pointset}}
\newcommand{\ctrldist}{\zeta}
\newcommand{\ctrlpta}{\point_{\bezlbl, 0}}
\newcommand{\ctrlptb}{\point_{\bezlbl, 1}}
\newcommand{\ctrlptc}{\point_{\bezlbl, 2}}
\newcommand{\ctrlptd}{\point_{\bezlbl, 3}}

\newcommand{\cost}{c}
\newcommand{\costpark}{\cost_{\parklbl}}
\newcommand{\costobs}{\cost_{\obslbl}}
\newcommand{\costt}{\cost_{\ts}}
\newcommand{\costsmooth}{\cost_{\smoothlbl}}
\newcommand{\costexp}{\cost_{\explbl}}
\newcommand{\costback}{\cost_{\backlbl}}

\maketitle


\begin{abstract}
In many applications of social navigation, existing works have shown that predicting and reasoning about human intentions can help robotic agents make safer and more socially acceptable decisions.
In this work, we study this problem for autonomous valet parking (AVP), where an autonomous vehicle ego agent must drop off its passengers, explore the parking lot, find a parking spot, negotiate for the spot with other vehicles, and park in the spot without human supervision.
Specifically, we propose an AVP pipeline that selects parking spots by explicitly predicting where other agents are going to park from their motion history using learned models and probabilistic belief maps.
To test this pipeline, we build a simulation environment with reactive agents and realistic modeling assumptions on the ego agent, such as occlusion-aware observations, and imperfect trajectory prediction.
Simulation experiments show that our proposed method outperforms existing works that infer intentions from future predicted motion or embed them implicitly in end-to-end models, yielding better results in prediction accuracy, social acceptance, and task completion.
Our key insight is that, in parking, where driving regulations are more lax, explicit intention prediction is crucial for reasoning about diverse and ambiguous long-term goals, which cannot be reliably inferred from short-term motion prediction alone, but can be effectively learned from motion history.
Website: \href{https://sites.google.com/view/chung2026selecting}{https://sites.google.com/view/chung2026selecting}.
\end{abstract}
\section{Introduction}

\begin{figure}[t]
\vspace*{1.7mm}
\centering
    \includegraphics[width=1\columnwidth]{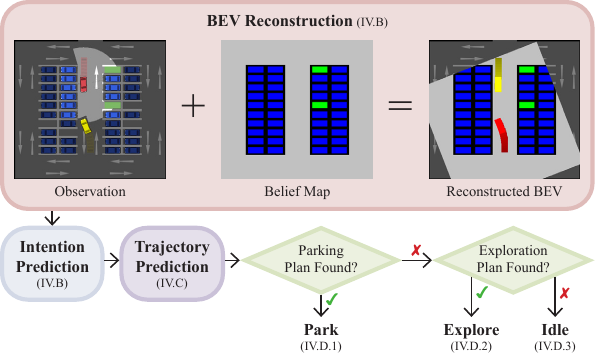}
\caption{
A flow chart of our method, with each component marked by its corresponding section number.
At each timestep, we combine information from the current observation and a probabilistic belief map of where cars are parked to reconstruct the bird's-eye view (BEV) map of other agents.
This enables the use of a learned intention model, which we use to make safe and socially acceptable parking decisions.
}\label{fig:pipieline}
\vspace*{-0.5cm}
\end{figure}

In social navigation, it is important for an autonomous ego agent to predict and reason with other human agents' \textit{intentions} (e.g.\ destination, choice of actions) in order to synthesize plans that are safe and socially acceptable.
Depending on the application (walking in crowds, driving through traffic, etc.), many existing works have proposed how and where intention prediction should fit within a planning framework.
For the specific problem of \textit{autonomous valet parking} (AVP), where a car must drop off its passengers, explore the parking lot, find a parking spot, negotiate for the spot with other vehicles, and park into the spot without human supervision, we define intentions as the intended parking spots of the other vehicles, the role of which we wish to investigate.
Specifically, we present an AVP pipeline that decides where an autonomous ego agent should park by explicitly predicting the intentions of other agents from motion history, which outperforms existing methods that infer intentions from trajectory predictions or handle them implicitly in an end-to-end fashion.
An overview of our method is shown in \Cref{fig:pipieline}.

\subsection{Related Work}
We now review how intentions are leveraged in other social navigation scenarios, and argue why they cannot be straightforwardly applied to the AVP problem.
We then review existing work on AVP and state our contribution.

\subsubsection{Intents in Social Navigation}\label{sec:traj_int_first}

Many existing works have proposed how and where intention predictions should fit within a planning framework in other applications of human-robot interactions.
For traffic scenarios such as lane merging, the driver's intention (whether it yields to ego or not) has been explicitly predicted from motion history using Probabilistic Graphical Model \cite{dong2017intention} or from trajectory predictions using heuristic-based \cite{schmidt2022meat} and learning-based \cite{park2023leveraging} intention models.
For crowd navigation, \cite{katyal2020intent} uses a generative adversarial network (GAN) to predict the intention (goal location) from past trajectories, whereas \cite{ratsamee2015social} inferred a pedestrian's intention (to avoid, maintain course, or approach the ego) from gaze directions and trajectory predictions.
There have also been many methods that implicitly handle other agents' intentions by mapping directly from observations to ego actions \cite{bojarski2016end, anjian_consistencyModel, codevilla2018end}.

\subsubsection{Autonomous Parking}
Unfortunately, results from \Cref{sec:traj_int_first} may not directly apply to parking, because intentions can be defined very differently in a parking lot: there are many spots where an agent may park, and vehicles do not necessarily follow lane directions and boundaries.
However, many existing works either address only parts of the AVP problem (e.g.\ exploration \cite{hu2023informative, lupien2024entropy}, deadlock resolution \cite{deadlockRecovery} and collision avoidance \cite{shen2023multi, shen2023reinforcement, shen2021collision}, generation of parking trajectories \cite{jiang2025hope, zhang2018autonomous, chi2023fast, nawaz2025graph}), or do not have humans in the loop \cite{shen2024parking}.
There is a significant lack of literature on spot selections and constructing a complete AVP pipeline; to the best of our knowledge, only \cite{nawaz2025occupancy} addresses this via an AVP pipeline where the ego selects spots with occupancy probability updates by inferring intentions from trajectory predictions.

That said, the results in \cite{nawaz2025occupancy} were obtained under certain simplifying assumptions: their sensing model ignores occlusion, simulated agents do not react to the ego's actions, and most crucially, future trajectories of all agents are assumed to be known.
Of course, there is no such thing as perfect trajectory prediction; but unlike other traffic scenarios, where good trajectory models can be trained from large amount of high quality motion data \cite{sun2020scalability, chang2019argoverse}, there is only one publicly available vehicle motion dataset for parking (to the best of our knowledge), the Dragon Lake Parking (DLP) Dataset \cite{shen2022parkpredict+}, which contains only 3.5 hours of footage from a single parking lot.

While \cite{shen2022parkpredict+, ji2024two} proposed trajectory prediction networks trained on this dataset, the trajectory predictions are conditioned on an intention prediction network.
As such, applying them to \cite{nawaz2025occupancy} would be \textit{paradoxical}: one would be predicting an intention from a trajectory predicted from an intention.
Moreover, these methods assume semantic bird's-eye view (BEV) information around the vehicle of interest, but BEV information is only realistically obtainable for the ego; predictions are usually needed for other agents, for which we only have partial BEV information.
In this paper, we propose an AVP pipeline that enables using \cite{shen2022parkpredict+} on other agents, which outperforms existing baselines when examined under more realistic modeling assumptions.

\subsection{Contributions}
Our core contribution is a new AVP pipeline that selects parking spots by explicitly predicting other agents' intention from motion history, which we find outperforms a variety of baselines.
We tested the method by constructing a simulation environment with realistic modeling choices, which include occlusion-aware sensing models, reactive agents, and is modular to the trajectory prediction method.
Our novel pipeline is enabled by three key supporting contributions:
\begin{enumerate}
\item\label{contri:avp_pipe}
We introduce a novel method to synthesize BEV information for surrounding agents using only information available to the ego, allowing intention models to predict where others are planning to park.
This enables the use of existing learned intention models for planning and decision-making in AVP.

\item We propose a parking spot selection strategy for the ego agent to avoid contended spots using predicted intentions.
We find that that this method performs better in prediction accuracy, social acceptance, and task completion compared to spot selection by explicitly inferring intents from trajectory predictions \cite{nawaz2025occupancy} and by implicit inferring intention in an end-to-end framework (using \cite{shen2022parkpredict+} on the ego only).

\item We propose an intention-conditioned trajectory prediction method for AVP using cubic B\'ezier curves.
Compared to other trajectory generation methods such as constant velocity, Hybrid A*, and the learned trajectory prediction model from \cite{shen2022parkpredict+}, our method best balances between low computation time, high prediction accuracy, and task completion rate.
\end{enumerate}
\section{Problem Formulation}\label{sec:prob}
In this paper, we aim to control the ego agent such that it successfully parks into a parking spot without colliding with other vehicles and pedestrians as soon as possible, while remaining within the parking lot's boundaries.
We now detail our assumptions on the parking lot and on the dynamics of vehicles and pedestrians, and formalize our problem statement at the end of this section.

\subsection{Parking Lot}\label{sec:lot_assum}
We denote the parking lot as $\lotset \subseteq \R^2$, which we assume is compact and connected.
We assume the parking lot has one entrance at coordinates $\point_{\entlbl} = \begin{bmatrix}\xs_{\entlbl} & \ys_{\entlbl}\end{bmatrix}\tp \in \lotset$, containing $\nspot \in \N$ rectangular parking spots $\spot_\is \subseteq \lotset$, $\is=1, \cdots, \nspot$, where each $\spot_\is$ is a rectangle parameterized by its length $\length_{\spot_\is} \in \Rpos$, width $\width_{\spot_\is}\in\Rpos$, 2-D coordinates of its center $\point_{\spot_\is} = \begin{bmatrix}\xs_{\spot_\is} & \ys_{\spot_\is}\end{bmatrix}\tp \in \lotset$, and heading in the direction that faces opposite of the road $\thetas_{\spot_\is} \in \R$.

Groups of parking spots are partitioned by $\nroad \in \N$ straight roads $\road_\js \subseteq \lotset$, $\js = 1, \cdots, \nroad$, on which vehicles can travel.
Each $\road_\js$ is a rectangle parameterized by a start point $\point_{\rstartlbl, \js} = \begin{bmatrix}\xs_{\rstartlbl, \js} & \ys_{\rstartlbl, \js}\end{bmatrix}\tp \in \R^2$, an end point $\point_{\rendlbl, \js} = \begin{bmatrix}\xs_{\rendlbl, \js} & \ys_{\rendlbl, \js}\end{bmatrix}\tp\in \R^2$ (which defines a center line $\roadcenter_{\js}$), and width $\width_{\road_\js} \in \Rpos$.
Mathematically, the rectangle is the Minkowski sum of $\roadcenter_{\js}$ and a square with side length $\width_{\road_\js}$, centered at the origin and rotated by $\thetas_{\road_\js} = \atantwo(\ys_{\rendlbl, \js} - \ys_{\rstartlbl, \js}, \xs_{\rendlbl, \js} - \xs_{\rstartlbl, \js})$.

We assume the ego agent has knowledge about the structure of the parking lot, specifically $\point_{\entlbl}$, $(\point_{\spot_\is})_{\is=1}^{\nspot}$, $(\thetas_{\spot_\is})_{\is=1}^{\nspot}$, $(\road_\js)_{\js=1}^{\nroad}$, and $\lotset$.
This assumption is not too restrictive and is also carried by many existing methods \cite{nawaz2025occupancy, shen2024parking, ji2024two}.
We leave generalization to unmapped parking lots as future work.

\subsection{Vehicle Model}
We model the vehicle agents as a Reeds-Shepp car in discrete time \cite{reeds1990optimal}, with dynamics
\begin{align}\label{eq:rs_car}
\begin{split}
    \state_{{\veh_\ks}, \ts + \dt} = \state_{{\veh_\ks}, \ts} + \begin{bmatrix}
        \vs_{{\veh_\ks}, \ts} \cos(\thetas_{{\veh_\ks}, \ts}) \\
        \vs_{{\veh_\ks}, \ts} \sin(\thetas_{{\veh_\ks}, \ts}) \\
        \omegas_{{\veh_\ks}, \ts},
    \end{bmatrix}\dt,
\end{split}
\end{align}
where $\ks \in \{0, 1, \cdots, \nother$\},
$\state_{{\veh_\ks}, \ts} = \begin{bmatrix}
    \point_{{\veh_\ks}, \ts}& \thetas_{{\veh_\ks}, \ts}
\end{bmatrix}\tp \in \R^2 \times [-\pi, \pi]$
is the state ($\point_{{\veh_\ks}, \ts} = \begin{bmatrix}\xs_{{\veh_\ks}, \ts} & \ys_{{\veh_\ks}, \ts}\end{bmatrix}\tp$ is the 2-D coordinates and $\thetas_{{\veh_\ks}, \ts}$ is the heading), $\ctrl_{{\veh_\ks}, \ts} = \begin{bmatrix}
    \vs_{{\veh_\ks}, \ts} & \omegas_{{\veh_\ks}, \ts}
\end{bmatrix}\tp \in \R^2$ is the control inputs of the vehicle ${\veh_\ks} \subseteq \R^2$ at time $\ts \in \{\taus\mid\mod(\taus, \dt)=0, \taus\in\R\}$, and $\dt \in \Rpos$ is the timestep size.

We denote by ${\veh_0}$ the ego vehicle and by ${\veh_1}, \cdots, {\veh_{\nother}}$ the other $\nother \in \{0\}\cup\N$ vehicle agents.
In particular, each $\veh_\ks$ is a rectangle $\{\begin{bmatrix}
    \xs & \ys
\end{bmatrix}\tp \mid -0.5\length_{\veh_\ks}\leq\xs\leq0.5\length_{\veh_\ks}, -0.5\width_{\veh_\ks}\leq\ys\leq0.5\width_{\veh_\ks}\}$ with length $\length_{\veh_\ks} \leq \min\{\length_{\spot_\is} \mid \is = 1, \cdots, \nspot\}$ and width $\width_{\veh_\ks} \leq \min\{\width_{\spot_\is} \mid \is = 1, \cdots, \nspot\}$.
At time $\ts$, the space each vehicle occupies $\veh_{\ks, \ts}$ is $\veh_\ks$ rotated by the heading $\thetas_{{\veh_\ks}, \ts}$ and translated by the coordinates $\point_{{\veh_\ks}, \ts}$.

\subsection{Pedestrian Model}
We model the pedestrians with holonomic dynamics:
\begin{align}\label{eq:ped_dyn}
    \point_{{\ped_\ms}, \ts + \dt} = \point_{{\ped_\ms}, \ts} + \begin{bmatrix}
        \vs_{{\ped_\ms}, \ts} \cos{(\thetas_{{\ped_\ms}, \ts})} \\
        \vs_{{\ped_\ms}, \ts} \sin{(\thetas_{{\ped_\ms}, \ts})}
    \end{bmatrix}\dt,
\end{align}
where $\ms \in \{1, \cdots, \nped\}$, $\nped\in\{0\}\cup\N$ is the number of pedestrians, $\point_{{\ped_\ms}, \ts} = \begin{bmatrix}
    \xs_{{\ped_\ms}, \ts} & \ys_{{\ped_\ms}, \ts}
\end{bmatrix}\tp \in \R^2$ is the 2-D position and $\ctrl_{{\ped_\ms}, \ts} = \begin{bmatrix}
    \vs_{{\ped_\ms}, \ts} & \thetas_{{\ped_\ms}, \ts}
\end{bmatrix}\tp \in \R^2$ is the control input of the pedestrian $\ped_\ms \subseteq \R^2$ at time $\ts$.

We assume the volume of each pedestrian $\ped_\ms$ is a disc with radius $\radius_{\ped_\ms}\in\Rnonneg$.
At time $\ts$, the space each pedestrian occupies $\ped_{\ms, \ts}$ is $\ped_\ms$ translated to $\point_{{\ped_\ms}, \ts}$.

\subsection{Problem Statement}
Mathematically, we wish to tackle the following problem:
\begin{problem}[AVP Problem]\label{prob:avp}
    Find $(\ctrl_{{\veh_0}, \ts})_{\ts=0}^{\tpark-\dt}$ and minimize the parking time $\tpark \in \{\taus\mid\taus\in\Rnonneg, \mod(\taus, \dt)=0\}$ such that $\exists\ \is \in \{1,\cdots,\nspot\}$, where
    \begin{align}
        {{\veh_{0, \tpark}}}\subseteq&{\spot_\is},\\
        {{\veh_{0, \ts}}} \cap {{\veh_{\ks, \ts}}} =& \emptyset,\\
        {{\veh_{0, \ts}}} \cap {{\ped_{\ms, \ts}}} =& \emptyset,\\
        {{\veh_{0, \ts}}} \subseteq & \lotset,
    \end{align}
    for all $\ts = 0, \cdots, \tpark$, $\ks=1, \cdots, \nother$, $\ms=1, \cdots, \nped$.
\end{problem}

In the next section, we detail the information available to the ego, and how each other vehicle and pedestrian agent reacts to the ego's actions at each timestep, which can make Problem \ref{prob:avp} non-trivial and more realistic.

\section{Simulation Environment}\label{sec:modeling}
We now detail our modeling choices on sensors and agents, with the goal of simulating occlusion-aware observations and reactive drivers and pedestrians in our environment.

\subsection{Observation Model}
To simulate an occlusion-aware sensing model, we employ a ray tracing algorithm.
To start, we assume we know a priori a compact, connected \textit{base sensing region} $\sensor_{0}\subseteq\R^2$ for the ego agent.
For example, $\sensor_0$ could be a rectangle or an ellipse.
At time $\ts = 0, \cdots, \tf-\dt$, where $\tf\in\{\taus\mid\taus\geq\tpark, \mod(\taus, \dt)=0\}$ is the final time, we define the \textit{sensing region} $\sensor_{0, \ts}\subseteq\R^2$ as $\sensor_{0}$ rotated by the ego's current heading $\thetas_{{\veh_0}, \ts}$ and translated by the coordinates $\point_{{\veh_0}, \ts}$.

At each $\ts$, the ego shoots out $\nray \in \N$ rays $\ray_{\js, \ts} \subseteq \sensor_{0, \ts}$, $\js=1, \cdots, \nray$, from the ego until it either hits the boundary of the parking lot, the sensing region, or another vehicle agent.
Mathematically, each $\ray_{\js, \ts}$ is the shortest line in a direction that begins at $\point_{{\veh_0}, \ts}$ and ends at a point in either $\partial\lotset$, $(\partial\veh_{\ks, \ts})_{\ks=1}^{\nother}$, or $\partial\sensor_{0, \ts}$, where $\partial\cdot$ refers to a set's boundary.
The ego is then able to observe all vehicles, spots, and pedestrians that intersect with any of the rays.

Furthermore, we assume the ego is able to distinguish between vacant spots ($\spot_\is\not\supseteq\veh_{\ks, \ts}\forall \ks\in\{1, \cdots, \nother\}$), occupied spots ($\spot_\is\supseteq\veh_{\ks, \ts}$ for some $\ks\in\{1, \cdots, \nother\}$), vehicles that are parked ($\veh_{\ks, \ts} \subseteq \spot_\is$ for some $\is\in\{1, \cdots, \nspot\}$, henceforth referred to as \textit{static vehicles}), and vehicles that are not parked ($\veh_{\ks, \ts} \not\subseteq \spot_\is\forall\is\in\{1, \cdots, \nspot\}$, henceforth referred to as \textit{dynamic vehicles}).
We denote the indices of observed vacant spots, occupied spots, static vehicles, dynamic vehicles, and pedestrians as $\spotset_{\vaclbl, \ts}, \spotset_{\occlbl, \ts}, \vehset_{\dynvehlbl, \ts}, \vehset_{\statvehlbl, \ts}, \pedset_{\dynpedlbl, \ts}\subseteq\N$ and their cardinality as $\ndim_{\vaclbl, \ts}, \ndim_{\occlbl, \ts}, \ndim_{\dynvehlbl, \ts}, \ndim_{\statvehlbl, \ts}, \ndim_{\dynpedlbl, \ts}\in\{0\}\cup\N$.

For each observed vehicle agent, the ego has access to their length $\length_{\veh_{\ks}}$, width $\width_{\veh_{\ks}}$, and current and some historical states for the past $\tpast$ time unit $(\state_{{\veh_\ks}, \taus})_{\taus=\ts - \tpast}^{\ts}\forall \ks\in\vehset_{\dynvehlbl, \ts}\cup\vehset_{\statvehlbl, \ts}$.
For each observed pedestrian, the ego has access to their radius $\radius_{\ped_{\ms}}$, and positions of their current and previous timestep $\point_{\ped_{\ms}, \ts-\dt}, \point_{\ped_{\ms}, \ts}\forall \ms \in \pedset_{\dynpedlbl, \ts}$.
The access to historical states is needed to enable the use of the learned intention model in \cite{shen2022parkpredict+}, as well as some of the trajectory prediction methods.

We assume the ego correctly observes every spots and agents it senses, even though real life sensors can have uncertainty about its observations, usually as a function of distance from the ego \cite{nawaz2025occupancy}.
We opt for this simpler model because it is unclear how uncertain, probabilistic observations can be straighforwardly used in the intention prediction model from \cite{shen2022parkpredict+}, which we leave as future work.

\subsection{Reactive Agents}

We aim to simulate other vehicles and pedestrians such that they would react to the ego's actions and not actively seek out collisions.
We accomplished this by introducing a parameter called \textit{passiveness} $\ndim_{\passivelbl, \agent} \in \{0\}\cup\N$, where $\agent\in\{\veh_1, \cdots, \veh_{\nother}, \ped_1, \cdots, \ped_{\nped}\}$ is another vehicle or pedestrian agent.
Intuitively, passiveness encodes how far into the future an agent begins braking.
Formally, it is the number of timesteps an agent looks ahead (based on its plan and other agents' current position) to check for collisions.
Higher passiveness makes an agent more cautious by considering collisions further ahead.
To implement this caution, our reactive agent brakes if it anticipates a collision, and resumes its plan otherwise; thus, a less passive agent is less likely to alter its behavior with stopping actions.

Mathematically, we assume each pedestrian and vehicle agent (except the ego) $\agent$ have some predefined plans $\ctrlplan_{{\agent}, \ts} \in \R^2$ for $\ts = -\tpast, \cdots, \tf + (\ndim_{\passivelbl, \agent}-1)\dt$, where $\tpast \in \{\taus\mid\taus\in\Rnonneg, \mod(\taus, \dt)=0\}$ is the time of some previous history.
At each timestep before $\ts = 0$, we control them by following the predefined plan (i.e.\ $\ctrl_{\agent, \ts}=\ctrlplan_{\agent, \ts}$ for $\ts = -\tpast, \cdots, -\dt$).
Afterwards, for each $\ts = 0, \cdots, \tf-\dt$, the agent would take the control $\ctrl_{\agent, \ts}=\zeros$ if $\agent_\taus \cap \veh_{0, \ts} \neq \emptyset$ for some $\taus=\ts+\dt, \cdots, \ts + \ndim_{\passivelbl, \agent}\dt$, or take the next unused plan from $(\ctrlplan_{{\agent}, \taus})_{\taus=0}^{\tf-\dt}$ otherwise.

While simple, this algorithm allows us to gauge the social acceptance of the ego agent by simulating spot competition, interruptions, and at-fault collisions, while also preventing the other agents from being self-destructive (as long as $\ndim_{\passivelbl, \agent} > 0$).
One could argue, however, that such heuristics may not necessarily reflect those of a real driver.
In contrast, prior works in social navigation have used data-driven trajectory models to better capture human behaviors \cite{shamsah2024socially}.
That said, we leave the development of more realistic reactive agents to future work, since existing trajectory models do not account for pedestrians and do not guarantee kinodynamically feasible trajectories \cite{shen2022parkpredict+}.
\section{Proposed Method: AVP Pipeline}

We now propose an AVP pipeline with explicit intention prediction from motion history.
Our overall plan is to enable the use of a learned intention prediction model such as \cite{shen2022parkpredict+} on the observed dynamic vehicles by reconstructing their BEV information using belief maps.
We then predict their future trajectories by conditioning cubic B\'ezier curves on the predicted intentions.
Finally, using Hybrid A* on the ego agent, we compute paths to park in spots less likely to be parked by other dynamic vehicles in the near future.
If no such path exists, the ego agent will instead choose to explore the parking lot for vacant spots.
An overview of our pipeline is shown in \Cref{fig:pipieline}.

\subsection{Intention Model}\label{sec:intent_pred}

Our goal is to predict the intentions of other observed dynamic vehicles (specifically, their probabilities of parking into each observed vacant spot, or to explore the parking lot by traveling to an \textit{exploration point}) using a learned intention model.
We opt for the model in \cite{shen2022parkpredict+}, as it is the only learned parking intention model with publicly available code, to the best of our knowledge.
Succinctly, at each time $\ts$ for each vehicle agent $\veh_\ks$, the intention model in \cite{shen2022parkpredict+} is a convolutional neural network (CNN) that outputs $\prob_{\ks, \is}\in[0, 1]$, the intention probability of a candidate parking spot $\spot_{\is}$, for each $\is$ in the index set $\spotset_{\canlbl, \ks}\subseteq\N$, and $\hat{\prob}_{\ks, \qs} \in [0, 1]$, the intention probability of a candidate exploration point $\state_{\explbl, \ks, \qs}\in\bigcup_{\js=1}^{\nroad}(\roadcenter_{\js}\times\{\thetas_{\road_\js}, \thetas_{\road_\js} + \pi\})$, for $\qs = 1$ to the number of candidate points $\ndim_{\explbl, \ks} \in \{0\}\cup\N$.
It takes as inputs the candidate exploration points $\state_{\explbl, \ks, \qs}$, the distance between the vehicle and the candidate spot $\dist_{\ks, \is}\in\Rnonneg$, the cosine similarity between the vehicle's heading vector and the vector from the vehicle to the candidate spot $\ang_{\ks, \is}\in[-1, 1]$, the predicted speed of the vehicle $\predvel_{{\veh_\ks}, \ts} \in \Rnonneg$, the distance between the vehicle and the parking lot entrance $\dist_{\entlbl}\in\Rnonneg$, the time the vehicle has spent in the parking lot $\ts_{\lotlbl}\in\{\taus\mid\taus\in\Rnonneg, \mod(\taus, \dt)=0\}$, and the semantic BEV image around the vehicle with the $\is^{\regtext{th}}$ spot marked $\img_{\ks, \is, \ts}\in\R^{400\times400\times3}$ (whereas $\img_{\ks, 0, \ts}$ do not have any spots marked differently).

Unfortunately, it is non-trivial to deploy this model in a realistic setting.
Specifically, from our modeling choices in \Cref{sec:modeling}, all required inputs are straightforward to obtain \textit{except} for the semantic BEV images $(\img_{\ks, \is, \ts})_{\is\in\{0\}\cup\spotset_{\canlbl, \ks}}$ for each observed vehicle $\veh_\ks$.
This is because the construction of these images requires knowledge of the occupied volume of all roads, parking spots (assumed known from \Cref{sec:lot_assum}), static vehicles, and current and past states (up to $\tpast$ time units) of dynamic vehicles within a 40 m square centered around the vehicle (of which we only observe a subset).
Though training a model without semantic BEV images is possible, ablation studies from \cite{shen2022parkpredict+} show significant performance loss.
\cite{shen2022parkpredict+} also only showed that their intention model improved performance in trajectory prediction, but did not suggest how it could be used for decision-making in AVP.
As such, we now propose a method to estimate and reconstruct the required BEV images with belief maps using only observed information, and suggest heuristics for spot selection based on this technique.

\subsection{BEV Reconstruction with Belief Map}\label{sec:belief_map}

Our strategy to reconstruct semantic BEV images for each observed dynamic vehicles $(\veh_\ks)_{\ks\in\vehset_{\dynvehlbl, \ts}}$ is to update a belief map of where cars are parked, and using that to infer the BEV of other agents.
We define the \textit{belief} $\belief_{\ts, \is} \in [0, 1]$ of a parking spot $\spot_\is$ at time $\ts$ as the estimated probability that other vehicles has parked or will park into the spot in the near future.
Since we know the number of spots $\nspot$ in the parking lot a priori, we store the ego's current belief of all parking spots into a \textit{belief map} $\beliefmap_{\ts} = \begin{bmatrix}
    \belief_{\ts, 1} & \cdots & \belief_{\ts, \nspot}
\end{bmatrix}\tp$ for each timestep.
We initialize $\beliefmap_{-\dt} = (0.5) \ones_{\nspot\times1}$ to show that we have no initial information about the occupancy probability of any of the parking spots.

At each timestep $\ts=0, \dt, \cdots$, we first update our belief with the current observation:
\begin{align}
    \intbelief_{\ts, \is} =
    \begin{cases}
        0 & \regtext{if}\ \is \in \spotset_{\vaclbl, \ts},\\
        1 & \regtext{if}\ \is \in \spotset_{\occlbl, \ts},\\
        \belief_{\ts-\dt, \is} & \regtext{otherwise},
    \end{cases}
\end{align}
for $\is=1, \cdots, \nspot$, where $\intbelief_{\ts, \is} \in [0, 1]$ is the initial belief.
In other words, we initially believe that each observed vacant and occupied spot will stay vacant or occupied for the near future.
For spots that we cannot observe, we keep our beliefs of them from the previous timestep.

We then estimate the BEV of each observed dynamic vehicles as follows: first, we draw the occupied volumes of the road, the spots (given from \Cref{sec:lot_assum}), and the current and past states of each observed vehicles (including the ego).
Then, for each unobserved spot $(\spot_\is)_{\is\notin\spotset_{\vaclbl, \ts}\cup\spotset_{\occlbl, \ts}}$ within the BEV, we draw a vehicle of some default length $\length_{\deflbl}$ and width $\width_{\deflbl}$ at the spot (centered at $\point_{\spot_\is}$ and rotated by $\thetas_{\spot_\is}$) if $\belief_{\ts, \is} \geq \beliefthresh$ (i.e.\ we believe it is \textit{occupied}) for some threshold $\beta\in[0, 1]$.

Finally, before the next timestep, we use the intention model in \Cref{sec:intent_pred} to update our belief.
For each observed dynamic vehicles, we define their candidate spots $(\spot_\is)_{\is\in\spotset_{\canlbl, \ks}}$ as all spots with $\belief_{\ts, \is} < \beliefthresh$ (i.e.\ we believe it is \textit{vacant}) within their BEV.
We then define their candidate exploration points and compute the other inputs to the intention model the same way as in \cite{shen2022parkpredict+}.
This enables us to compute the intention probabilities $\prob_{\ks, \is}$ (and $\hat{\prob}_{\ks, \qs}$) of each observed dynamic vehicle parking into each observed vacant spot, which we can use to update the belief as:
\begin{align}
\begin{split}
    &\belief_{\ts, \is} =\\
    &\begin{cases}
        1 - \prod_{\prob\in\{\prob_{\tilde{\ks}, \tilde{\is}}\mid\tilde{\is}=\is\in\spotset_{\canlbl, \tilde{\ks}}, \tilde{\ks} \in \vehset_{\dynvehlbl, \ts} \}}(1-\prob) & \regtext{if}\ \is \in \spotset_{\vaclbl, \ts},\\
        \intbelief_{\ts, \is} & \regtext{otherwise}.
    \end{cases}
\end{split}
\end{align}
In other words, we update the belief of each observed vacant spot as the probability that \textit{any} of the observed dynamic vehicle will park into the spot in the near future.

\subsection{Trajectory Prediction}\label{sec:traj_pred}
To maximize the utility of our method in \Cref{sec:belief_map}, we now condition trajectory predictions of other agents on their predicted intentions for collision avoidance.
At each timestep $\ts$, for each observed dynamic vehicle $(\veh_\ks)_{\ks\in\vehset_{\dynvehlbl, \ts}}$, we wish to predict trajectories for all intentions with probability higher than some threshold $\goalthresh\in[0, 1]$.
That is, for each $\prob_{\ks,\is}\geq\goalthresh$ and $\hat{\prob}_{\ks, \qs}\geq\goalthresh$, we wish to predict $(\state_{\veh_{\ks}, \taus})_{\taus=\ts+\dt}^{\ts+\tpred}$ with predictions $(\predstate_{\veh_{\ks}, \taus})_{\taus=\ts+\dt}^{\ts+\tpred}$, where $\predstate_{\veh_{\ks}, \taus} \in \R^2\times[-\pi, \pi]$ is the prediction of the ground truth $\state_{\veh_{\ks}, \taus}$ and $\tpred\in\{\taus\mid\taus\in\Rnonneg, \mod(\taus, \dt)=0\}$ is the prediction time.

To accomplish this, we employ a goal-conditioned trajectory prediction scheme using cubic B\'ezier curves.
A cubic B\'ezier curve is a 2-D spline defined by two control points $\ctrlpta, \ctrlptd \in \R^2$ that define the start and end points, and two control points $\ctrlptb, \ctrlptc \in \R^2$ that determines the slope at the beginning and the end of the curves \cite{mortenson1999mathematics}.
To connect the agent to the intention with the appropriate end points and slopes, we define the control points as:
\begin{align}
    \ctrlpta &= \point_{\veh_{\ks}, \ts},\\
    \ctrlptd &= \point_{\goallbl},\\
    \ctrlptb &= \point_{\veh_{\ks}, \ts} + \zeta\predvel_{{\veh_{\ks}}, \ts}\begin{bmatrix}
        \cos{(\thetas_{\veh_{\ks}, \ts} + \pi)}\\
        \sin{(\thetas_{\veh_{\ks}, \ts} + \pi)}
    \end{bmatrix},\\
    \ctrlptc &= \point_{\spot_\is} + \zeta\predvel_{{\veh_{\ks}}, \ts}\begin{bmatrix}
        \cos{(\thetas_{\goallbl} + \pi)}\\
        \sin{(\thetas_{\goallbl} + \pi)}
    \end{bmatrix},
\end{align}
where the goal is the center and heading of the spot if the intention is to park ($\point_{\goallbl}=\point_{\spot_\is}$, $\thetas_{\goallbl} = \thetas_{\spot_\is}$) or as the exploration point if the intention is to explore ($\begin{bmatrix}
    \point_{\goallbl} & \thetas_{\goallbl}
\end{bmatrix} = \state_{\explbl, \ks, \qs}$), and $\zeta\in\Rnonneg$ adjusts the distance between the end control points and the control points in-between.

We can now predict the dynamic vehicles' trajectories.
For $\taus = \ts+\dt, \cdots, \ts+\tpred$, we compute $\predstate_{\veh_{\ks}, \taus}$ by assuming that the vehicle travels at a constant velocity $\predvel_{{\veh_{\ks}}, \ts}$ along the curve.
In the case where $\tplan - \tpred$ more timesteps of predictions are needed for some $\tplan \in \{\taus\mid\taus>\tpred, \mod(\taus, \dt)=0\}$, we fill the rest of the predicted timesteps using a constant velocity model.

Note that we do not predict the intentions of the pedestrians due to the lack of intention models for pedestrians in a parking lot.
Instead, for each observed pedestrians $(\ped_\ms)_{\ms\in\pedset_{\dynpedlbl, \ts}}$, we only use a constant velocity model to predict their trajectories $(\predstate_{\ped_{\ms}, \taus})_{\taus=\ts+\dt}^{\ts+\max\{\tpred, \tplan\}}$, where $\predstate_{\ped_{\ks}, \taus} \in \R^3$ is the prediction of the ground truth $\state_{\ped_{\ks}, \taus}$.
We predict the velocity of the pedestrian as $\predvel_{{\ped_\ms}, \ts} = \frac{\norm{\point_{{\ped_\ms}, \ts} - \point_{{\ped_\ms}, \ts - \dt}}_2}{\dt}$.

\subsection{Final Trajectory Planning}
After the intention and trajectory prediction in \Cref{sec:intent_pred,sec:belief_map}, the ego carries out its decisions similar to the pipeline in \cite{nawaz2025occupancy} at each timestep $\ts$, which we succinctly summarize here:

\begin{figure}[t]
\vspace*{1.7mm}
\centering
    \includegraphics[width=1\columnwidth]{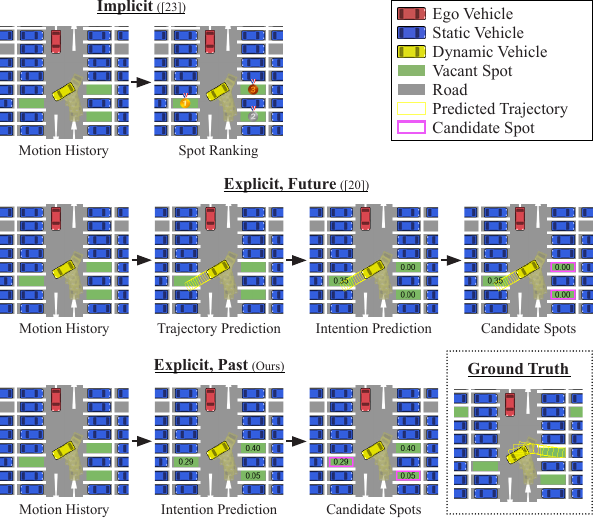}
\caption{
An overview of the methods under comparison, with legend shown on top right.
In this scenario, the dynamic vehicle (other agent) is backing into a spot after moving forward to adjust its angle (a ``right-south-up'' maneuver \cite{shen2024parking}).
Our proposed method is the only one to explicitly and correctly predicted this behavior.
}
\label{fig:traj_vs_int}
\vspace*{-0.5cm}
\end{figure}

\subsubsection{Parking}\label{sec:parking}
Our ego first considers parking if a safe path to a non-contended spot is found.
We define the ego's candidate parking spots $(\spot_\is)_{\is\in\spotset_{\canlbl, 0}}$ as all observed vacant spots with belief less than some threshold $\egothresh \in [0, 1]$, i.e.\ $\spotset_{\canlbl, 0} = \{\is \mid \belief_{\ts, \is} \leq \egothresh, \is\in\spotset_{\vaclbl, \ts}\}$.

For each of the candidate spot, we use the Hybrid A* implementation from \cite{nawaz2025occupancy} to compute $(\ctrl_{{\veh_0}, \taus})_{\taus=\ts}^{\ts + \tplan - \dt}$ such that our ego agent is parked by the end of the plan while avoiding spots we believed to be occupied.
That is, ${\veh_{0, \ts+\tplan}}\subseteq\spot_\is$ and ${\veh_{0, \taus}} \subseteq  \lotset$, ${\veh_{0, \taus}} \cap \{\spot_{\tilde{\is}} \mid \belief_{\taus, \tilde{\is}} \geq \beliefthresh, \tilde{\is}=\{1, \cdots, \nspot\}\} = \emptyset$ for $\taus = \ts + \dt, \cdots, \ts + \tplan$.

If this path also avoids collisions with the predicted trajectories of vehicles and pedestrians, the ego executes the plan until obstructed by obstacles from new observations and predictions.
In the case of multiple feasible plans, the ego agent would choose the path with the lowest cost $\costpark \in \R$, computed as in \cite{nawaz2025occupancy}.

\subsubsection{Exploration}\label{sec:explore}
If no safe parking plan is found, our ego considers exploring the parking lot at the exploration points $(\state_{\explbl, 0, \qs})_{\qs=1}^{\ndim_{\explbl, 0}}$, where $\state_{\explbl, 0, \qs}=\begin{bmatrix}
    \point_{\explbl, 0, \qs} & \thetas_{\explbl, 0, \qs}
\end{bmatrix}\tp \in \bigcup_{\js=1}^{\nroad}(\roadcenter_{\js}\times\{\thetas_{\road_\js}, \thetas_{\road_\js} + \pi\})$, are defined as the intersection points between the center lines of the road $(\roadcenter_{\js})_{\js=1}^{\nroad}$ and the boundary of the sensing region $\partial\sensor_{0, \ts}$.

We compute and select the lowest-cost paths with Hybrid A* similar to \Cref{sec:parking}.
In addition, we condition that our ego to first consider exploration points in front ($\begin{bmatrix}
    \cos(\thetas_{\veh_0, \ts})&
    \sin(\thetas_{\veh_0, \ts})
\end{bmatrix}(\point_{\explbl, 0, \qs} - \point_{\veh_0, \ts})\geq0$) before considering those behind ($\begin{bmatrix}
    \cos(\thetas_{\veh_0, \ts})&
    \sin(\thetas_{\veh_0, \ts})
\end{bmatrix}(\point_{\explbl, 0, \qs} - \point_{\veh_0, \ts})<0$).
In practice, the ego would only consider exploring backwards if the road ahead of it stays blocked for the foreseeable future.

\subsubsection{Idling}\label{sec:idle}
If no safe plans were found in both \Cref{sec:parking,sec:explore}, the ego would stay put at the current position for this timestep (i.e.\ $\ctrl_{\veh_0, \ts}=\zeros$) to wait for the other agents to move out of the ego's way. 
\section{Experiments}\label{sec:results}
We now compare our method, which predicts intention from motion history, with methods that infer intentions from trajectory predictions or implicitly in an end-to-end framework, as summarized in \Cref{fig:traj_vs_int}.
We also evaluate alternative trajectory prediction methods against our cubic Bézier baseline from \Cref{sec:traj_pred}. 
All experiments were performed using Python on a laptop computer with a 24-core i9 CPU, 64 GB RAM, and an NVIDIA RTX 4090 GPU.

\begin{figure}[t]
\vspace*{1.7mm}
\centering
    \includegraphics[width=1\columnwidth]{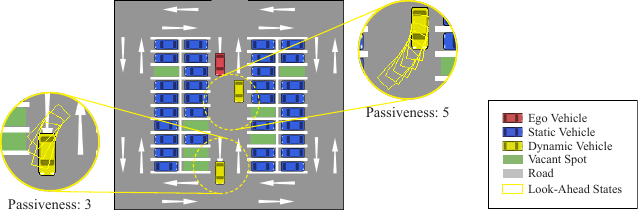}
\caption{
Setup of our experiment at $\ts=0$. 
At each iteration, some of the bottom 10 spots are randomly chosen to be vacant.
Then, 1 or 2 dynamic vehicles would spawn, with pots being their goal.
There are always 2 vacant spots on the sides in case the ego failed to compete.
}\label{fig:legend}
\vspace*{-0.5cm}
\end{figure}

\subsection{Environment Setup}\label{sec:env_setup}

We conduct our experiments in a small parking lot with four columns of 10 spots, divided by 3 vertical and 2 horizontal roads.
At each iteration, between 1 and 10 of the bottom 10 spots in the 2 middle columns are vacant, depending on whether 1 or 2 dynamic vehicles are present.
2 additional spots on the outer columns are also vacant in case the ego fails to compete for the bottom spots.
The ego agent always starts at the top of the middle vertical road, facing downward.
We illustrate this setup in \Cref{fig:legend}.

At the start of each iteration, 1 or 2 dynamic vehicles spawn, each randomly assigned a vacant spot at the bottom.
Their spawn locations and parking maneuvers are chosen randomly from the 8 maneuvers defined in \cite{shen2024parking}, which are characterized by whether the vehicle is in the closer or the further lane, whether it starts before or after the spot, and whether it parks head-in or tail-in (see \cite[Figure 4]{shen2024parking}).
If these predefined plans collide with each other, a new target spot and a parking maneuver are randomly chosen again.

All scenarios were ran with the hyperparameters $\tf=100 \regtext{ s}$, $\sensor_0 = \{\point\mid\norm{\point}_2 \leq 11.5 \regtext{ m}, \point\in\R^2\}$, $\nray=360$, $\tpast=4\regtext{ s}$, $\length_{\deflbl} = 4.97\regtext{ m}$, $\width_{\deflbl} = 1.86\regtext{ m}$, $\beliefthresh=0.5$, $\goalthresh=0.3$, $\zeta=3\regtext{ s}$, and $\egothresh=0.3$.
Any unmentioned variables were chosen to match those in \cite{nawaz2025occupancy} for fair comparison.
Note that all compared methods only differs in how they choose the ego’s parking spot; exploration, pedestrian interaction, sensing, and other components were kept constant.
As such, the experiments were conducted in a small parking lot with no pedestrians to best observe the effects of the different methods.

\subsection{Comparison Setup}
To gauge the methods' ability to predict and avoid collision, we randomized 500 different environment setups as in \Cref{sec:env_setup}, and simulate them with reactive agents with passiveness chosen uniformly from 2 to 6.
To assess how the methods perform under more challenging scenarios, we also ran simulations in the same 500 setups with non-reactive vehicle agents (i.e.\ $\ndim_{\passivelbl, \veh_1} = \cdots = \ndim_{\passivelbl, \veh_{\nother}} = 0$).

We compare the methods under a variety of evaluation metrics.
To measure task completion, we recorded success rate, where a task is considered unsuccessful if collision occurred or the ego failed to arrive at a goal before $\tf$.
To measure social acceptance, we recorded the number of iterations where the ego has ``stolen'' a spot originally targeted by a dynamic vehicle, as well as the average number of timesteps interrupting the dynamic agents' predefined plans.
Since all dynamic agents were initialized closer to their intended spots, any takeover by the ego represents a true case of spot stealing.
To measure the quality of the resulting trajectories, we recorded $\tpark$, the average time the ego takes to park into a goal.
To measure computation efficiency, we recorded the average time it took for an ego to select the goal at each timestep (since each method differs in how they select spots), and the average time it took to plan for a trajectory after selecting the goal (which we expect to be the same).
Finally, to measure the accuracy of the trajectory prediction, we recorded the average minimum average displacement error (minADE) and the minimum final displacement error (minFDE) \cite{chang2019argoverse} of each episode.

We compared our method against different trajectory conditioning methods, \cite{nawaz2025occupancy}, and an implementation of \cite{shen2022parkpredict+}.
The implementation details are as follows:

\subsubsection{Effects of Trajectory Prediction Method}
\begin{figure}[t]
\vspace*{1.7mm}
\centering
    \includegraphics[width=1\columnwidth]{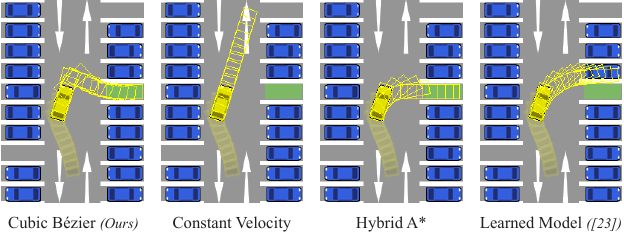}
\caption{
Example trajectory predictions (yellow outline) conditioned on the predicted intention (green spot) which we compared in our experiment.
}\label{fig:traj_pred}
\vspace*{-0.5cm}
\end{figure}

To validate our choice of using cubic B\'ezier curve in \Cref{sec:traj_pred}, we replaced it with a constant velocity model, with Hybrid A*, and with the learned trajectory prediction model from \cite{shen2022parkpredict+} (a transformer-based model) to predict the trajectories of each other agent conditioned on the predicted intent location.
Some example trajectories are shown in \Cref{fig:traj_pred}.

\subsubsection{Effects of Inferring Intentions from Trajectory Prediction}\label{sec:traj_first}
We compared against \cite{nawaz2025occupancy} to investigate the effect of inferring intentions from the future vs.\ the past and to evaluate its performance under more realistic assumptions.
Due to the lack of non-intention-driven trajectory prediction methods for parking agents, we replaced their ground truth trajectory prediction with a constant velocity model.

\subsubsection{Effects of Implicit Intent Reasoning}
To investigate whether explicit reasoning of intentions are beneficial, we implemented \cite{shen2022parkpredict+} by applying the intention prediction model on \textit{the ego only}.
At each timestep, we apply the intention prediction model on the ego, and constant velocity trajectory prediction on other agents (for the same reason as \Cref{sec:traj_first}).
The ego would then prioritize spots with the highest intention probability that it can find a safe plan to.

\subsection{Results and Discussion}
The results of the experiments are shown in \Cref{table:results}.
For non-reactive agents, our proposed method has a significantly higher success rate ($\approx$+9\%) than \cite{nawaz2025occupancy} and \cite{shen2022parkpredict+}, possibly due to our better accuracy in trajectory prediction, as seen from the lower average minADE ($\approx$-1 m) and minFDE ($\approx$-3 m).
Since our environment and agent behavior were not from the DLP dataset \cite{shen2022parkpredict+}, this shows that the intention prediction model generalizes to our setup.

For reactive agents, since agents no longer actively seek out collision, our method's improvement in success rate against \cite{nawaz2025occupancy} and \cite{shen2022parkpredict+} is less ($\approx$+1\%), but we perform better in social acceptance, stealing less spots ($\approx$-4\%) and interrupting the plans of the other agents less ($\approx$-0.2), which we attribute to the improved accuracy in trajectory prediction.

Of the four trajectory prediction methods in our proposed AVP pipeline, intention-conditioned cubic B\'ezier curve consistently perform better in task completion, social acceptance, and computation time across both non-reactive and reactive agents.
Surprisingly, it generated trajectories with lower minADE and minFDE than those from constant velocity and Hybrid A*, and with comparable results with the learned model in \cite{shen2022parkpredict+}, which may explain its better performance.

In general, all methods produce trajectories of similar quality when comparing their average parking time.
That said, our proposed method suffers from a higher computation time in spot selection ($\approx$+0.005 s) compared to \cite{nawaz2025occupancy} and \cite{shen2022parkpredict+}, likely due to our need to reconstruct the semantic BEV map and predict others' intents with neural network inference. 
However, this increase is quite insignificant compared to the time needed for generating the paths, which takes $\approx$1-1.5 s to compute.
While such planning times fall short of the 10 Hz cycle typically desired for deployment, our pipeline is modular and can readily incorporate faster trajectory planners: whether learning-based \cite{jiang2025hope}, optimization-based \cite{zhang2018autonomous}, reachability-based \cite{chi2023fast}, or sampling-based \cite{nawaz2025graph}.
Since our study centers on the role of intention modeling and spot selection, we leave planner comparisons and computation optimizations to future work.

\begin{table*}[!ht]
\centering
\captionsetup{font=small}
\small
\caption{Results of our proposed method with different trajectory prediction methods, a method that predicts intentions from trajectory prediction \cite{nawaz2025occupancy}, and a method that reasons with intentions implicitly \cite{shen2022parkpredict+} for 500 parking simulations with non-interactive agents and 500 simulations with interactive agents.
The best results in each evaluation metric ($\uparrow$: higher is better, $\downarrow$: lower is better) are highlighted.}
\label{table:results}
\begin{tabular}{l|l||r||r|r||r|r||r||r|r}
     \multicolumn{2}{c||}{} & \multicolumn{1}{c||}{Success} & \multicolumn{2}{c||}{Social Acceptance} & \multicolumn{2}{c||}{Computation Time (s) $\downarrow$} & \multicolumn{1}{c||}{Parking} & \multicolumn{2}{c}{Accuracy}\\
     \cline{4-7}\cline{9-10}
     \multicolumn{2}{c||}{Method} & \multicolumn{1}{c||}{Rate} & \multicolumn{1}{c|}{Spots} & \multicolumn{1}{c||}{Interrupted} & \multicolumn{1}{c|}{Spot} & \multicolumn{1}{c||}{Path} & \multicolumn{1}{c||}{Time} & \multicolumn{1}{c|}{minADE} & \multicolumn{1}{c}{minFDE}\\
     \multicolumn{2}{c||}{} & \multicolumn{1}{c||}{(\%) $\uparrow$} & \multicolumn{1}{c|}{Stolen (\%) $\downarrow$} & \multicolumn{1}{c||}{Timesteps $\downarrow$} & \multicolumn{1}{c|}{Selection} & \multicolumn{1}{c||}{Planning} & \multicolumn{1}{c||}{(s) $\downarrow$} & \multicolumn{1}{c|}{(m) $\downarrow$} & \multicolumn{1}{c}{(m) $\downarrow$}\\
     \hline\hline \multicolumn{10}{c}{Non-Reactive Agents}\\
     \hline\hline  & Cubic & \multirow{2}{*}{\textbf{99.8}} & \multirow{2}{*}{\textbf{0.0}} & \textbf{0.000} & 0.0052 & \textbf{1.2812} & 20.1 & 2.02 & \textbf{2.40}\\
& B\'ezier (Ours) & & & $\pm$\textbf{0.000} & $\pm$0.0026 & $\pm$\textbf{0.2516} & $\pm$9.6 & $\pm$1.22 & $\pm$\textbf{1.89}\\
\cline{2-10} & Constant & \multirow{2}{*}{92.0} & \multirow{2}{*}{0.2} & \textbf{0.000} & 0.0058 & 1.2848 & \textbf{19.3} & 3.09 & 5.59\\
Explicit, & Velocity & & & $\pm$\textbf{0.000} & $\pm$0.0031 & $\pm$0.2479 & $\pm$\textbf{9.2} & $\pm$1.33 & $\pm$2.32\\
\cline{2-10} Past & Hybrid & \multirow{2}{*}{94.6} & \multirow{2}{*}{\textbf{0.0}} & \textbf{0.000} & 0.7360 & 1.3435 & 20.1 & 2.46 & 3.81\\
& A* & & & $\pm$\textbf{0.000} & $\pm$0.3257 & $\pm$0.2825 & $\pm$9.7 & $\pm$1.39 & $\pm$2.04\\
\cline{2-10} & Learned & \multirow{2}{*}{96.4} & \multirow{2}{*}{0.2} & \textbf{0.000} & 0.0535 & 1.3476 & 20.9 & \textbf{1.82} & 2.46\\
& Model (\cite{shen2022parkpredict+}) & & & $\pm$\textbf{0.000} & $\pm$0.0278 & $\pm$0.2717 & $\pm$10.5 & $\pm$\textbf{1.16} & $\pm$1.63\\
\hline \multicolumn{2}{c||}{Implicit} & \multirow{2}{*}{90.4} & \multirow{2}{*}{0.2} & \textbf{0.000} & \textbf{0.0006} & 2.9143 & 21.1 & 2.77 & 5.00\\
\multicolumn{2}{c||}{(\cite{shen2022parkpredict+})} & & & $\pm$\textbf{0.000} & $\pm$\textbf{0.0000} & $\pm$1.5890 & $\pm$8.8 & $\pm$1.30 & $\pm$2.30\\
\hline \multicolumn{2}{c||}{Explicit, Future} & \multirow{2}{*}{90.8} & \multirow{2}{*}{\textbf{0.0}} & \textbf{0.000} & 0.0010 & 1.5907 & 19.4 & 3.13 & 5.64\\
\multicolumn{2}{c||}{(\cite{nawaz2025occupancy})} & & & $\pm$\textbf{0.000} & $\pm$0.0001 & $\pm$0.4023 & $\pm$9.4 & $\pm$1.35 & $\pm$2.33\\
     \hline\hline \multicolumn{10}{c}{Reactive Agents}\\
     \hline\hline  & Cubic & \multirow{2}{*}{\textbf{100.0}} & \multirow{2}{*}{\textbf{0.0}} & \textbf{0.072} & 0.0052 & \textbf{0.7564} & 20.4 & 2.02 & \textbf{2.39}\\
& B\'ezier (Ours) & & & $\pm$\textbf{0.599} & $\pm$0.0027 & $\pm$\textbf{0.1698} & $\pm$9.4 & $\pm$1.22 & $\pm$\textbf{1.88}\\
\cline{2-10} & Constant & \multirow{2}{*}{99.0} & \multirow{2}{*}{2.8} & 0.870 & 0.0057 & 0.7621 & \textbf{19.2} & 2.98 & 5.38\\
Explicit, & Velocity & & & $\pm$8.316 & $\pm$0.0031 & $\pm$0.1796 & $\pm$\textbf{8.8} & $\pm$1.23 & $\pm$2.18\\
\cline{2-10} Past & Hybrid & \multirow{2}{*}{98.4} & \multirow{2}{*}{2.8} & 0.672 & 0.4274 & 0.7966 & 20.0 & 2.39 & 3.75\\
& A* & & & $\pm$8.216 & $\pm$0.2041 & $\pm$0.1907 & $\pm$9.1 & $\pm$1.28 & $\pm$1.92\\
\cline{2-10} & Learned & \multirow{2}{*}{99.0} & \multirow{2}{*}{4.4} & 0.824 & 0.0529 & 0.8062 & 20.7 & \textbf{1.80} & 2.46\\
& Model (\cite{shen2022parkpredict+}) & & & $\pm$11.032 & $\pm$0.0266 & $\pm$0.2205 & $\pm$10.1 & $\pm$\textbf{1.16} & $\pm$1.64\\
\hline \multicolumn{2}{c||}{Implicit} & \multirow{2}{*}{98.6} & \multirow{2}{*}{3.4} & 0.208 & \textbf{0.0006} & 1.6834 & 20.9 & 2.52 & 4.59\\
\multicolumn{2}{c||}{(\cite{shen2022parkpredict+})} & & & $\pm$1.022 & $\pm$\textbf{0.0000} & $\pm$0.8825 & $\pm$8.5 & $\pm$1.01 & $\pm$1.86\\
\hline \multicolumn{2}{c||}{Explicit, Future} & \multirow{2}{*}{98.8} & \multirow{2}{*}{4.6} & 0.408 & 0.0010 & 1.0238 & \textbf{19.2} & 2.90 & 5.28\\
\multicolumn{2}{c||}{(\cite{nawaz2025occupancy})} & & & $\pm$7.334 & $\pm$0.0001 & $\pm$0.3210 & $\pm$\textbf{9.0} & $\pm$1.16 & $\pm$2.06
     
\end{tabular}
\end{table*}
\section{Conclusion}\label{sec:conclusion}

This work investigated the role of intention prediction on the AVP problem.
By proposing a new AVP pipeline and conducting extensive experiments in a realistic simulation environment, we conclude that explicit intention prediction from motion history is beneficial to selecting spots in a safer, more socially acceptable manner compared to intention prediction from trajectory models and implicit, end-to-end reasoning of intentions.
We argue that this is because explicit reasoning of intentions are necessary to handle the diverse and ambiguous long-term goals in parking, but the maneuvers executed to achieve these intentions are often too complex to be reliably inferred from short-term cues (such as using constant velocity models) alone.
We speculate that this finding may also be applicable to robotic social navigation with similar conditions, such as service robots operating in shared indoor spaces, or autonomous delivery systems with docking stations in public space.

\subsubsection*{Limitations}
We observed three limitations of our method.
Firstly, our current implementation does not yet achieve real-time performance, as Hybrid A* dominates computation for path planning.
However, because the pipeline is modular, it could readily incorporate faster planners in future work.
Secondly, our experiments were performed on simulations with predefined agent paths rather than real humans, which may limit the ecological validity of our findings.
While we are preparing human-in-the-loop simulations and real-vehicle experiments, they must be preceded by achieving real-time computation.
Finally, our intention model does not yet reason with pedestrians, vehicles exiting from their spots, or agent types such as bicycles, motorcycles, and trucks.
This limitation is carried over from the literature that we use for comparison \cite{shen2022parkpredict+}.
Towards this, we are working on constructing and training our own model from the DLP dataset \cite{shen2022parkpredict+}.





\renewcommand{\bibfont}{\normalfont\footnotesize}
{\renewcommand{\markboth}[2]{}
\printbibliography}

\end{document}